\documentclass[11pt]{article}

\usepackage[final]{acl}

\usepackage{times}
\usepackage{latexsym}
\usepackage[T1]{fontenc}
\usepackage[utf8]{inputenc}
\usepackage{microtype}
\usepackage{inconsolata}
\usepackage{graphicx}
\usepackage{booktabs}
\usepackage{multirow}
\usepackage{amsmath}
\usepackage{xcolor}
\usepackage{tikz}
\usetikzlibrary{shapes.geometric, arrows.meta, positioning}

\title{Iterating Toward Better Search:\\A Two-Agent Simulation Framework for\\Evaluating Agentic Search Architectures in E-Commerce}

\author{
  Jetlir Duraj$^{*}$,
  Jayanth Yetukuri,
  Shuang Zhou,
  Dhruv Varma,
  Rui Kong,
  Ishita Khan,
  Qunzhi Zhou \\
  eBay Inc. \\
  \texttt{\{jduraj, jyetukuri, shuazhou, dvarma, rukong, ishikhan, qunzhou\} @ebay.com} \\[0.3em]
  {\small $^{*}$Corresponding author.}
}

\begin{document}
\maketitle


\begin{abstract}
We present a modular two-agent simulation framework for evaluating conversational
shopping assistant architectures. An independent buyer agent, configured with
personas, missions, and patience levels, is paired with an interchangeable
responder that integrates with a real e-commerce search API. Holding the buyer
constant across experiments enables controlled comparison of responder designs on
identical scenarios. Using 2{,}011 conversations across 14 persona buckets, we
establish four empirical findings. First, rolling-window memory outperforms
intent-extraction memory on all quality metrics while being 35\% faster per query.
Second, illustrating rapid evidence-driven iteration, a systematic failure analysis of a responder version
enables targeted fixes that reduce failure and near-failure rates by 62\% across the full dataset. Third,
swapping the responder LLM backbone from Gemini~2.5 to Llama~3.3~70B costs 0.16--0.45
points despite identical architecture. Finally, we document that judge selection is itself a consequential architectural
decision: SOTA Gemini and Claude models disagree on 30\% of conversations by two or more points
despite identical prompts.
\end{abstract}


\section{Introduction}
\label{sec:intro}

Building conversational shopping assistants that reliably serve diverse buyer needs
  requires controlled evaluation of architectural choices, memory strategies,
  intent extraction, response generation, before committing to production
  deployment. In an ideal world, one would conduct AB tests at
  scale covering all relevant buyer types with enough statistical power to draw
  conclusions. In practice, high-powered AB tests covering a buyer
  typology are rarely achievable at low cost. When introducing a new
  technology such as conversational shopping to a customer base, it is difficult to
  anticipate all the ways buyer behavior will evolve once the technology is deployed.
  Cheap, reproducible offline evaluation tools that cover diverse buyer types
  become essential for system iteration and improvement.
  Once the framework is in place, marginal cost per iteration is near zero, given LLM access.

Two standard approaches fall short. \textit{Beta user testing} provides realistic
signals but is slow, cannot replay identical scenarios across candidate
architectures, and raises privacy constraints. \textit{Single-agent generation},
where one model produces both buyer queries and assistant responses, is faster but
produces conversations that differ systematically from real user behavior: queries
tend to be more formal, action
commands (clicking or putting in a cart) are rarer, and the generated buyer ``knows'' what the
assistant will return before asking. Critically, single-agent generation cannot
test \textit{responder-specific} architectural features,
because these components do not participate in the generation at all.

We present a \textbf{modular two-agent simulation} system that addresses both
limitations. A \textit{buyer agent} generates queries independently, reacting only
to what the responder actually returns; a \textit{responder agent} processes those
queries using a real e-commerce search API and returns search result pages (SRP)
and/or conversational guidance (CHAT). Because either agent can be replaced without
modifying the other, architectures can be compared on identical buyer scenarios
with the buyer held as a controlled variable. The buyer's configurability also
allows systematic testing of how buyer behavior may evolve after launch.

Using this framework with 2{,}011 buyer conversations across 14 persona buckets, we
conduct four experiments illustrating the iterative cycle of responder architecture
design and evaluation. Our contributions are:
\begin{itemize}
  \item A modular two-agent simulation system enabling controlled, reproducible
    comparison of responder architectures (\S\ref{sec:system}).
  \item An empirical demonstration that simpler rolling-window memory may outperform
    explicit intent extraction while reducing per-query latency (\S\ref{sec:v5Sys-B}).
  \item A systematic study of failures from 2{,}011 conversations that directly
    enables targeted fixes of a responder architecture,  recovering $+$1.3--1.8~points on the worst-performing
    scenarios and reducing near-failures by 62\% across the full corpus, validated in just 2 days (\S\ref{sec:Sys-BSys-B2}).
  \item Evidence that underlying LLM contributes independently of architecture:
    Llama~3.3~70B vs Gemini~2.5 costs 0.16 - 0.45~points (\S\ref{sec:Sys-BSys-C}).
  \item Evidence that frontier LLM judges embed different evaluation philosophies
    from an identical prompt: exact agreement on CHAT Helpfulness is only 13\%,
    30\% of conversations show $\geq$2-point divergence, and the evaluator gap
    ($\approx$0.5~points) exceeds the architecture gap, making judge selection a
    consequential design decision (\S\ref{sec:judges}).
\end{itemize}


\section{Related Work}
\label{sec:related}

\noindent\textbf{Behavioral fidelity evaluation.}
\citet{wang2025opera} and \citet{lu2025can} evaluate shopping agents by measuring
how closely predicted next actions match historical user logs, treating real human
sessions as the ground truth. This optimises for \textit{mimicry} rather than
system effectiveness: a responder that predicts the next click correctly need not
be one that actually helps users accomplish their goals. Our framework moves the
metric towards mission success, and buyer behavior is generated fresh each run rather than replayed from a log.



\noindent\textbf{Production monitoring.}
  \citet{zhao2025agent} embed human feedback into a live retrieval flywheel to prevent
  knowledge decay; \citet{warne2026simulation} use simulation to stress-test prompt
  changes against a structured ``case state'' reasoning architecture before deployment.
  Both treat an architecture as fixed and optimise its operation. Our
  framework is an \textit{R\&D laboratory}: it tests \textit{architectures}
  before commitment, mapping failure boundaries that production flywheels may not be able to
  surface.

\noindent\textbf{Personalization and satisfaction evaluation.}
\citet{zhao2025personalens} and \citet{sun2025llm} use LLM agents to score
personalization or satisfaction as aggregate metrics. We instead evaluate
\textit{architectural design choices} and expose that the evaluator is itself a
design choice (\S\ref{sec:judges}).

\noindent\textbf{Memory architectures for conversational agents.}
\citet{shinn2023reflexion} introduce Reflexion, showing episodic memory with
self-reflection improves agent performance. \citet{packer2023memgpt} demonstrate
memory compression can maintain performance while reducing context length. We
empirically validate that simpler rolling-window memory can outperform intent-extraction
pipelines in conversational commerce.

\noindent\textbf{Agent architectures and tool use.} \citet{schick2023toolformer} show models can learn to use external APIs through demonstration; \citet{yao2023react} propose ReAct, interleaving reasoning and action for improved tool use. Our responder architectures integrate tool use with real search APIs, and our empirical comparison confirms that simpler designs can outperform more complex extraction pipelines.

\noindent\textbf{LLM-as-judge.}
Using LLMs to evaluate model outputs is now standard practice
\citep{zheng2023judging}. \citet{dubois2024alpacafarm} show that LLM judges can
match human agreement on many tasks, while \citet{chiang2023can} document
systematic biases in model-based evaluation. We extend this paradigm with a
systematic cross-judge comparison on identical shopping conversations, showing that
judge selection is equivalent to selecting a quality criterion and that the resulting
gap may be large enough to affect architectural
conclusions.


\section{System Architecture}
\label{sec:system}

\subsection{Two-Agent Design}

\begin{figure}[t]
\centering
\begin{tikzpicture}[
    node distance=1.5cm,
    box/.style={rectangle, rounded corners=3pt, draw=black, thick,
                minimum width=5.5cm, minimum height=1.3cm,
                align=center, font=\small},
    arr/.style={-{Stealth[length=5pt]}, thick},
    lbl/.style={font=\scriptsize\itshape}]

  \node[box, fill=blue!8] (buyer) {%
    \textbf{Buyer Agent}\\[3pt]
    {\scriptsize configurable: mission, persona, tone, patience}};

  \node[box, fill=gray!12, below=1.5cm of buyer] (orch) {%
    \textbf{Orchestrator}\\[3pt]
    {\scriptsize routes messages \& logs all turns \& conversation stats}};

  \node[box, fill=orange!10, below=1.5cm of orch] (resp) {%
    \textbf{Responder Agent}\\[3pt]
    {\scriptsize memory, classifiers, search API, responses DB}};

  \draw[arr] ([xshift=-8pt]buyer.south) --
             node[left, lbl, pos=0.5]{query}
             ([xshift=-8pt]orch.north);
  \draw[arr] ([xshift=-8pt]orch.south) --
             node[left, lbl, pos=0.5]{query}
             ([xshift=-8pt]resp.north);

  \draw[arr] ([xshift=8pt]resp.north) --
             node[right, lbl, pos=0.5]{\{SRP, CHAT\}}
             ([xshift=8pt]orch.south);
  \draw[arr] ([xshift=8pt]orch.north) --
             node[right, lbl, pos=0.5]{\{SRP, CHAT\}}
             ([xshift=8pt]buyer.south);

\end{tikzpicture}
\caption{Two-agent simulation. Either agent is replaceable, enabling controlled
architecture comparisons.}
\label{fig:architecture}
\end{figure}
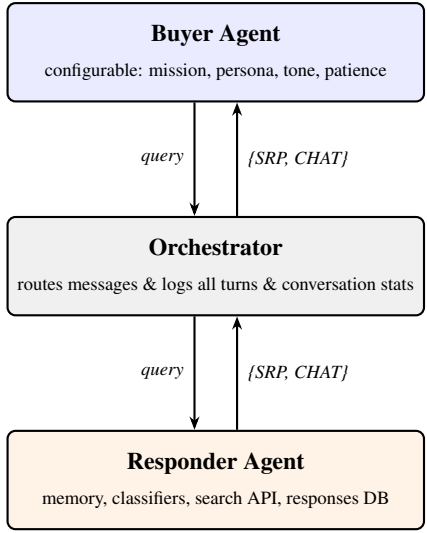

Figure~\ref{fig:architecture} shows the system we build. The \textbf{orchestrator} runs a
conversation loop: it forwards buyer queries to the responder, returns the response
to the buyer, logs the turns and various dynamic conversation statistics, and repeats until the buyer sends \texttt{[TERMINATE\_SESSION]} or
exhausts its turn budget. The critical design property is \textit{agent
independence}: the buyer observes only the responder's output (SRP and/or CHAT);
the responder observes only the buyer's raw text query; and either component can be
replaced
without modifying the other. This enables controlled experiments where
only the responder architecture changes across runs.

\subsection{Buyer Agent}
\label{sec:buyer}

\begin{figure}[t]
\centering
\scalebox{0.8}{%
\begin{tikzpicture}[
    node distance=0.45cm,
    proc/.style={rectangle, rounded corners=2pt, draw, thick, fill=blue!7,
                 minimum width=4cm, minimum height=1cm,
                 align=center, font=\scriptsize},
    dec/.style={diamond, draw, thick, fill=orange!12, aspect=2.6,
                align=center, font=\scriptsize, inner sep=1pt},
    term/.style={rectangle, rounded corners=8pt, draw, thick, fill=gray!15,
                 minimum width=2.0cm, minimum height=0.5cm,
                 align=center, font=\scriptsize},
    act/.style={rectangle, rounded corners=2pt, draw, thick, fill=green!10,
                minimum width=1.5cm, minimum height=0.45cm,
                align=center, font=\tiny},
    arr/.style={-{Stealth[length=4pt]}, thick},
    lbl/.style={font=\tiny\itshape}]

  \node[proc] (ctx)  {Prepare context\\(mission + memory)};
  \node[proc, above=of ctx] (chk) {Check termination / turn budget};
  \node[term, above=of chk] (start) {New turn};
  \node[term, right=0.5cm of chk, font=\tiny, minimum width=1.1cm] (end1) {TERMINATE};
  \node[proc, below=of ctx] (pmt)  {Build LLM prompt};
  \node[proc, below=of pmt] (llm)  {LLM call (Gemini 2.5 Pro)};
  \node[dec,  below=of llm] (type) {Action\\type?};

  \node[act, below left=0.55cm and 0.45cm of type]  (srch) {Search query};
  \node[act, below=0.55cm of type]                   (clk)  {Click item};
  \node[act, below right=0.55cm and 0.45cm of type]  (cart) {Cart add};

  \node[dec, below=0.5cm of cart] (val) {itemId in\\last SRP?};
  \node[term, below=0.35cm of val, minimum width=1.2cm, font=\tiny] (rej) {Discard};

  \node[proc, below=2.5cm of type, minimum width=1.6cm] (mem) {Update memory\\+ compact\\if needed};
  \node[term, below=of mem]        (ret) {Return query string};

  \draw[arr] (start) -- (chk);
  \draw[arr] (chk)   -- node[lbl, left]{ok} (ctx);
  \draw[arr] (chk)   -- node[lbl, above]{limit} (end1);
  \draw[arr] (ctx)   -- (pmt);
  \draw[arr] (pmt)   -- (llm);
  \draw[arr] (llm)   -- (type);

  \draw[arr] (type) -| node[lbl, above left,  pos=0.05]{search} (srch);
  \draw[arr] (type) -- node[lbl, right]{click}                  (clk);
  \draw[arr] (type) -| node[lbl, above right, pos=0.05]{cart}   (cart);

  \draw[arr] (cart)    --                              (val);
  \draw[arr] (val.south) -- node[lbl, right]{no}      (rej.north);

  \draw[arr] (srch.south) |- (mem.west);
  \draw[arr] (clk.south)  -- (mem.north);
  \draw[arr] (val.west)   -| node[lbl, above, pos=0.25]{yes} (mem.north);

  \draw[arr] (mem) -- (ret);

  \draw[arr] (ret.west) -- ++(-1.8,0)
    -- node[lbl, left, pos=0.5]{\{SRP, CHAT\}} ++(0,9.5)
    -- ([xshift=-1.8cm]start.west) -- (start.west);
\end{tikzpicture}}%
\caption{Buyer Agent query generation loop. Each turn: check budget, prepare context
from three memory streams (queries/SRPs/CHATs), call LLM, produce one action type.
Cart additions are validated against the current SRP.}
\label{fig:buyer_workflow}
\end{figure}
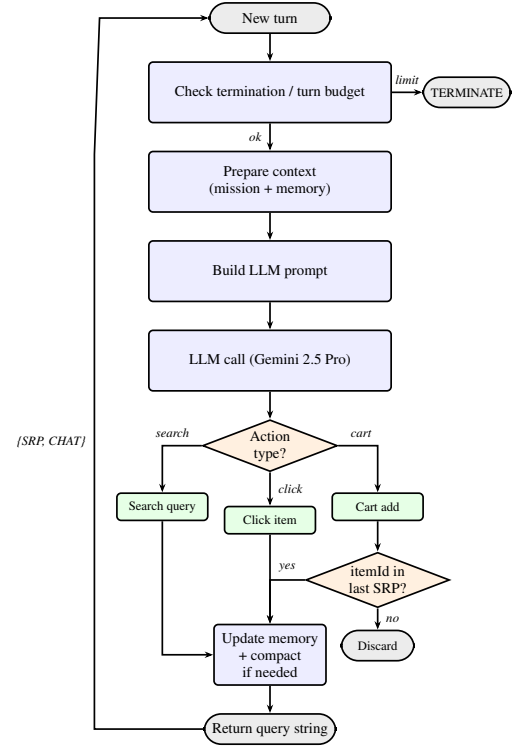

The buyer is a Gemini~2.5~Pro reasoning model operating in a mission-based loop
(Figure~\ref{fig:buyer_workflow}).\footnote{We also experimented with OpenAI
GPT-5.2 and GPT-5.4 as buyer LLMs but found that the Gemini~2.5 series generates
more realistic buyer speech patterns.} Each session contains 1--3 independent
missions;
each mission specifies a shopping goal, buyer persona, communication tone, and
patience level encoded as \texttt{max\_turns} (4~turns for impatient buyers;
10~turns for patient buyers). The buyer persona, patience and communication tone do not vary across missions of the same buyer session. The buyer LLM and the set of buyer configurations is held \textit{constant} across all
experiments. Hence, performance differences are due to the responder
architecture alone.

\paragraph{Memory architecture.}
The buyer maintains three parallel streams: \textbf{query memory} (recent queries,
auto-compacted when $>$10); \textbf{SRP memory} (keeps most recent 3 when
$>$5); \textbf{CHAT memory} (keeps most recent 3 when $>$5). On mission transition,
the last 3 turns are injected into the next mission's context, enabling natural
handoffs (e.g., ``ok changing topic'') rather than abrupt resets.

\paragraph{Query generation pipeline.}
Each turn, the buyer (1)~checks whether the current mission has terminated or
exhausted its turn budget; (2)~prepares context from the current mission goal,
persona/tone, and all three memory streams; (3)~constructs an LLM prompt with
available actions and cart command syntax; (4)~calls Gemini~2.5~Pro (temperature
0.2); (5)~parses and validates the response; and
(6)~updates memory, triggering compaction if any threshold is exceeded.
\vspace{-1mm}
\paragraph{Action types and cart logic.}
The buyer produces three action types: \textbf{search query} (natural language or
keyword-style), \textbf{item click} (requests full listing details),
and \textbf{cart addition} (primary purchase-intent signal). Cart additions are validated against current SRP
results only, preventing hallucinated actions. A mission terminates on satisfaction,
turn-budget exhaustion, or explicit abandonment.

\subsection{Responder Architectures}
\label{sec:responders}

All responders share the same pipeline structure: a \textbf{memory module}, an
\textbf{orchestration layer via query understanding classifiers and query re-writers}, \textbf{real search API integration} returning
live eBay listings, and a \textbf{response generator}. They differ in memory
design and underlying LLM.

\paragraph{Sys-A: Intent Tracking.}
Memory stores raw queries \textit{and} LLM-extracted intent statements. After each
buyer query, a dedicated LLM call extracts structured intents and generates a
condensed search digest, adding one extra LLM call per query.

\paragraph{Sys-B: Rolling Window.}
Memory accumulates raw buyer queries; once the count exceeds 6, all accumulated
queries are compacted into a single keyword digest via one LLM call and accumulation
restarts.\footnote{A systematic ablation over the compaction threshold; rerunning the same 2{,}011 conversations with thresholds of 4, 6, 8, and 10 is left for future work.} There is no per-query intent-extraction call (unlike Sys-A); the digest
LLM call fires only at compaction, reducing per-query latency by 35\% compared to Sys-A.

\paragraph{Sys-B+: Targeted Fixes of Sys-B.}
3 targeted changes addressing failure patterns,  \S\ref{sec:fulleval}.

\paragraph{Sys-C: Llama Backbone.}
Same rolling-window architecture and orchestration as Sys-B; the generative backbone is
changed from Gemini~2.5~Pro/Flash to Llama~3.3~70B~Instruct.


\section{Experimental Setup}
\label{sec:setup}

\subsection{Dataset}
\label{sec:dataset}

We use a dataset of 2{,}011 buyer conversation configurations spanning 14~buckets
defined by three axes: \textit{shopping style} (information-seeking,
broad-to-narrow, precise-flexible, precise-strict), \textit{patience}
(patient: 10~turns per mission; impatient: 4~turns), and \textit{mission count}
(1--3~missions per session). Table~\ref{tab:dataset} shows the full distribution.
Seed keyword queries are drawn from eBay consumer search logs spanning all verticals, supplemented by thematic and occasion-based queries; an LLM then expands each seed into a full buyer mission specifying goal, context, constraints, and persona. The 14-bucket structure was designed to cover buyer archetypes relevant to agentic search evaluation. The dataset is intended to illustrate the framework; teams evaluating specific architectures will construct datasets targeting their own verticals, time windows, and query patterns, which the modular framework readily supports. 

\begin{table}[t]
\centering
\small
\setlength{\tabcolsep}{4pt}
\begin{tabular}{p{3.2cm}lrr}
\toprule
\textbf{Style} & \textbf{Patience} & \textbf{Miss.} & \textbf{$N$} \\
\midrule
Information seeking\\(no purchase intent)
                    & Patient           & 1 & 200 \\
\midrule
\multirow{3}{*}{\parbox{3.2cm}{Broad to narrow\\(explores,\\narrows by results)}}
                    & \multirow{3}{*}{Patient}   & 1 & 160 \\
                    &                   & 2 & 200 \\
                    &                   & 3 & 100 \\
\midrule
\multirow{6}{*}{\parbox{3.2cm}{Precise but flexible\\(open to alternatives)}}
                    & \multirow{3}{*}{Patient}   & 1 & 160 \\
                    &                   & 2 & 200 \\
                    &                   & 3 & 100 \\
\cmidrule{2-4}
                    & \multirow{3}{*}{Impatient} & 1 & 133 \\
                    &                   & 2 & 200 \\
                    &                   & 3 & 100 \\
\midrule
\multirow{4}{*}{\parbox{3.2cm}{Precise and strict\\(exact match,\\no alternatives)}}
                    & Patient            & 1 &  25 \\
\cmidrule{2-4}
                    & \multirow{3}{*}{Impatient} & 1 & 133 \\
                    &                   & 2 & 200 \\
                    &                   & 3 & 100 \\
\midrule
\multicolumn{3}{l}{\textbf{Total}} & \textbf{2{,}011} \\
\bottomrule
\end{tabular}
\caption{Dataset distribution across 14 buckets. Each bucket is defined by shopping/browsing
style, patience level, and mission count. Five combinations are excluded as
less relevant (e.g., multi-mission patient strict). Patient buyers receive
10~turns/mission; impatient buyers only 4.}
\label{tab:dataset}
\end{table}

The following is an example of a buyer configuration from the \texttt{precise\_flexible\_patient\_1mission} bucket.

\begin{quote}\small
\textbf{Mission:} ``Need original Apple Watch sports loop band, but willing to
consider alternatives if they offer better value or are in excellent condition.''\\[2pt]
\textbf{Patience:} \texttt{max\_turns = 10} (patient)\\
\textbf{Persona:} Experienced buyer who knows what they want but recognises good
alternatives when presented; professional but open-minded.\\
\textbf{Tone:} Casual conversational; asks follow-up questions, appreciates
suggestions.
\end{quote}

\subsection{Evaluation}
\label{sec:eval}

We use two different frontier LLM judges to evaluate conversations along four dimensions (1--5 scale), with
chain-of-thought reasoning required before scoring.\footnote{Traditional IR metrics such as NDCG or precision@k are not applicable here for two structural reasons. First, the buyer reacts to each responder's output, so the same buyer configuration generates different queries against different systems, breaking the fixed-query assumption IR metrics depend on. Second, even at temperature~0, LLM inference is not fully deterministic due to floating-point non-associativity in GPU kernels \citep{atil2025nondeterminism}, making relevance annotations from one run non-transferable to the next.} This dual-judge design provides
robustness and as we show in \S\ref{sec:judges} exposes systematic differences in evaluation philosophy.

\paragraph{Evaluation metrics.}
Each conversation is scored on four metrics:
\begin{itemize}
  \item \textbf{Mission Success}: Did the buyer accomplish their stated goal?
    Measures outcome quality based on buyer actions and satisfaction signals
    (5: buyer clicks or carts an item, or expresses strong purchase intent;
    3: buyer makes progress but does not find fully satisfactory results;
    1: no meaningful progress).
  \item \textbf{SRP Relevance}: How relevant were search results to buyer needs?
    Evaluates query-to-search translation and appropriateness of SHOW decisions
    (5: all results match query and accumulated constraints, SHOW decisions always
    appropriate; 3: 50\%+ results match or some inappropriate SHOW decisions;
    1: $<$20\% relevance or critical failure).
  \item \textbf{CHAT Helpfulness}: How helpful was conversational guidance?
    Did responses leverage search data to provide specific, actionable
    information and were CHAT decisions appropriate?
    (5: grounded guidance with specific product or policy information, CHAT decisions
    always appropriate; 3: somewhat helpful but generic, or occasional inappropriate
    decisions; 1: unhelpful or consistently inappropriate CHAT usage).
  \item \textbf{Query Intent Understanding}: Did the responder correctly interpret
    each query and activate the right pipeline components across all turns? (5: correct routing and constraint maintenance throughout; 3: main intent
    correct but important context lost from memory or some classification errors;
    1: systematic mis-classification or complete failure to maintain conversational
    context.)
\end{itemize}

\paragraph{LLM judges.}
We employ two frontier models, both receiving identical evaluation prompts: \\
\hspace*{1em} i) \textbf{Gemini~3.1~Pro} \citep{reid2026gemini} \\
  \hspace*{1em} ii) \textbf{Claude~Opus~4.6} \citep{anthropic2026claude} \\
We aggregate by taking the mean across metrics. When both judges agree
directionally (e.g., Sys-B $>$ Sys-A), findings are robust; substantial disagreement
indicates architectural trade-offs (\S\ref{sec:judges}) rather than noise.
The evaluation prompt and a sample of over 200 conversations spanning the score range were manually reviewed by the authors to confirm that judge scores align with human assessment of interaction quality; low-scoring conversations were verified to represent failed interactions and high-scoring ones to represent successful journeys.

\paragraph{Reproducibility.}
The buyer agent workflow (Figure~\ref{fig:buyer_workflow}) and evaluation metrics
are described in this paper. The responder architectures are described at 
the level of design decisions (\S\ref{sec:responders}). Implementation code for
responder agents is proprietary and cannot be released. The buyer configurations and
conversation transcripts are similarly not available for external release. The eBay search API is an internal service and is not publicly accessible. We make
these constraints explicit in the interest of transparency; limited reproducibility
is a standard constraint for many industry papers.


\section{Results}
\label{sec:results}

\subsection{Architecture Comparison: Sys-A vs.\ Sys-B}
\label{sec:v5Sys-B}

We run the buyer configurations across all 14 buckets; each run through both Sys-A and Sys-B.

\paragraph{Outcome: simpler memory wins.}
Contrary to the initial hypothesis, Sys-B outperforms Sys-A on every metric under both
judges (Table~\ref{tab:v5Sys-B}). The advantage is modest (0.01--0.10 points) but
consistent across all metrics and judge conditions. In head-to-head per-conversation
comparisons, Sys-B wins more often than Sys-A under both judges: Gemini (Sys-B: 23.2\%,
Sys-A: 15.2\%, tie: 61.6\%) and Claude (Sys-B: 27.3\%, Sys-A: 26.3\%, tie: 46.5\%). Sys-B is also 35\% faster per query. 

\begin{table}[t]
\centering
\small
\setlength{\tabcolsep}{5pt}
\begin{tabular}{lr}
\toprule
\textbf{Metric} & \textbf{$\Delta$ (Sys-B$-$Sys-A)} \\
\midrule
\multicolumn{2}{l}{\textit{Gemini-3.1 Pro judge}} \\
Mission Success      & $+$0.10 \\
SRP Relevance        & $+$0.07 \\
CHAT Helpfulness     & $+$0.08 \\
Intent Understanding & $+$0.05 \\
\midrule
\multicolumn{2}{l}{\textit{Claude Opus 4.6 judge}} \\
Mission Success      & $+$0.01 \\
\midrule
\multicolumn{2}{l}{\textit{Speed}} \\
Latency (per query)  & $-$35\% \\
\bottomrule
\end{tabular}
\caption{Sys-A (intent tracking) vs.\ Sys-B (rolling window).
Sys-B wins all metrics under both judges while being 35\% faster per query. All
Claude scores align with Gemini direction; mission success is representative.}
\label{tab:v5Sys-B}
\end{table}

\paragraph{Why simpler memory wins.}
Intent extraction introduces failure modes absent in Sys-B: the dedicated LLM call can
miscategorise intents, injecting noise downstream, and the compressed intent representation can drop information present in the raw query. 
Modern long-context reasoning models appear
capable of working directly with raw query sequences, making intermediate extraction
a liability in most conversation lengths.
Sys-A occasionally wins on very long conversations ($\geq$3~missions, patient
buyers), suggesting explicit intent tracking may help when queries span 15+~turns
and important information is missed by Sys-B's short 6-query window, but these cases are rare (<5\% of dataset).

\subsection{Evaluation of Sys-B}
\label{sec:fulleval}

\paragraph{Overall performance.}
Sys-B exhibits strong aggregate performance
(Table~\ref{tab:Sys-Bfull}). Gemini scores consistently exceed Claude across all
metrics; the gap between judges (0.39--0.97~points) reflects different evaluation
philosophies rather than disagreement about architecture quality (\S\ref{sec:judges}).

\begin{table}[t]
\centering
\resizebox{\columnwidth}{!}{%
\begin{tabular}{lrrr}
\toprule
\textbf{Metric} & \textbf{Gemini (\%)} & \textbf{Claude (\%)} & \textbf{Gap (pts)} \\
\midrule
Mission Success      & 90.4 & 81.4 & $+$0.46 \\
SRP Relevance        & 87.0 & 77.4 & $+$0.49 \\
CHAT Helpfulness     & 92.2 & 73.0 & \textbf{$+$0.97} \\
Intent Understanding & 89.8 & 82.0 & $+$0.39 \\
\bottomrule
\end{tabular}}
\caption{Sys-B evaluation. Mean scores as percentage of maximum (5/5), by judge,
and the absolute Gemini--Claude gap. The CHAT Helpfulness gap (0.97~pts) is the
largest, reflecting Sys-B's tendency to generate informative explanations that
Gemini rewards but Claude deems insufficiently action-oriented.}
\label{tab:Sys-Bfull}
\end{table}

\paragraph{Persona analysis.}
Sys-B excels with patient, flexible buyers: information-seeking and broad-to-narrow
exploratory buckets rank highest, with multi-mission patient configurations among
the strongest performers. The rolling-window architecture is well-matched to buyers who allow time for discovery. The worst-performing combination is \texttt{precise\_strict\_impatient}: all distinct
mission-number variants rank in the bottom three buckets. 

\paragraph{Failure analysis.}
We define catastrophic failure as all four metrics scored 1/5. Of 2{,}011
conversations, 14 (0.7\%) meet this criterion---but failures are highly
concentrated. Seven occur in the single
\texttt{precise\_strict\_impatient\_1mission} bucket ($N$=133), a 5.3\% rate that
is 7.5$\times$ the overall average. All patient buckets with $N\geq$100 achieve
0.0\% catastrophic failure, confirming that patient buyers give Sys-B time to recover
from missteps. Examining the broader set of 65 low-scoring conversations (score
$\leq$2/5 on all metrics) a few recurring failure patterns emerge. These provide actionable targets for
the next iteration (\S\ref{sec:Sys-BSys-B2}).

\paragraph{Perfect success.}
838 conversations (41.7\%) score 5/5 on all metrics under Gemini. Only 96 (4.8\%)
achieve this under both judges simultaneously. The 36.9-percentage-point gap
quantifies Sys-B's architectural trade-off: strong at process quality and
conversational engagement (Gemini's criterion); inconsistent at converting engagement into concrete buyer outcomes (Claude's criterion).

\subsection{Rapid Iteration: Sys-B $\rightarrow$ Sys-B+}
\label{sec:Sys-BSys-B2}

The failure analysis in \S\ref{sec:fulleval} directly motivated targeted changes in Sys-B+, identified and implemented in one working day. Recurring patterns in Sys-B's degraded responses informed three targeted architectural changes addressing the most frequent failure modes.

\paragraph{Evaluation of Sys-B+ }                
  We first validate on Sys-B's 29 worst-performing configurations---the                                                                                                                                                                                
  \texttt{precise\_strict\_impatient} bucket conversations scoring $\leq$2.0 (Claude
  average). Table~\ref{tab:Sys-BSys-B2} shows consistent improvement across all three bucket                                                                                                                                                              
  variants and both judges; with the targeted fixes, buyers adapt their search strategy
  and in many cases complete their missions. Running Sys-B+ across all 2{,}011                                                                                                                                                                 
  conversations (Table~\ref{tab:Sys-BSys-B2full}), both judges agree on direction: Gemini                                                                                                                                                                 
  overall average rises by $+$1.6\%; Claude by $+$1.0\%. CHAT Helpfulness shows the
  largest per-metric gain ($+$2.2\% for both judges). Failure rates improve sharply: catastrophic failures fall from 14                    
  to 9~($-$36\%); near-failures from 65 to 25~($-$62\%); Gemini-perfect conversations               
  rise from 41.7\% to 46.1\%. Gains concentrate in \texttt{psi} buckets---where exact                    
  constraints most often produce 0-result responses and impatient buyers have fewest turns to recover. One \texttt{psi} bucket (\texttt{precise\_strict\_impatient\_2missions}) reaches significance under Gemini ($p$$=$$0.004$, Welch's $t$-test), and a sign test across all 14~bucket deltas gives $p$$=$$0.029$. 

\begin{table}[!ht]
\centering
\small
\setlength{\tabcolsep}{4.5pt}
\begin{tabular}{lrr}
\toprule
\textbf{Bucket} & \textbf{Claude $\Delta$} & \textbf{Gemini $\Delta$} \\
\midrule
\texttt{psi-1m} ($n$=17) & $+$1.3 & $+$1.7 \\
\texttt{psi-2m} ($n$=10) & $+$1.3 & $+$1.8 \\
\texttt{psi-3m} ($n$=2)  & $+$1.4 & $+$1.8 \\
\midrule
\textbf{Overall} ($n$=29) & $+$\textbf{1.3} & $+$\textbf{1.8} \\
\bottomrule
\end{tabular}
\caption{Sys-B+ vs.\ Sys-B score improvement on 29 worst-performing
\texttt{precise\_strict\_impatient} (\texttt{psi}) configurations. Targeted fixes
recover $+$1.3 (Claude) and $+$1.8 (Gemini) points consistently across all three
bucket variants.}
\label{tab:Sys-BSys-B2}
\end{table}




\begin{table}[t]
\centering
\small
\begin{tabular}{lr}
\toprule
& \textbf{$\Delta$ (Sys-B+ vs.\ Sys-B)} \\
\midrule
\multicolumn{2}{l}{\textit{Mean score change, 2{,}011 conversations}} \\
Gemini overall avg      & $+$1.6\% \\
\quad CHAT Helpfulness  & $+$2.2\% \\
Claude overall avg      & $+$1.0\% \\
\quad CHAT Helpfulness  & $+$2.2\% \\
\midrule
\multicolumn{2}{l}{\textit{Failure and perfection rates}} \\
Catastrophic (all 1/5)   & $-$36\% \\
Near-failure ($\leq$2/5) & $-$62\% \\
Gemini-perfect (all 5/5) & $+$4.4pp \\
\bottomrule
\end{tabular}
\caption{Sys-B+ vs.\ Sys-B full-scale evaluation, all 2{,}011 conversations. Score
change shown as percentage of baseline. Targeted fixes reduce near-failures by 62\%
and catastrophic failures by 36\%. Gains concentrate in \texttt{psi} buckets; the
2-mission variant reaches significance ($p$$=$$0.004$, Gemini). A sign test across
all 14 bucket deltas gives $p$$=$$0.029$.}
\label{tab:Sys-BSys-B2full}
\end{table}


\subsection{LLM Quality: Sys-B vs.\ Sys-C}
\label{sec:Sys-BSys-C}

Having established that \textit{architecture} affects quality (Sys-A vs.\ Sys-B), we
also verify how the \textit{underlying LLM} contributes independently. Sys-C holds
Sys-B's architecture constant and swaps the generative backbone from
Gemini~2.5~Pro/Flash to vLLM-served Llama~3.3~70B~Instruct.

\paragraph{Results.}
Table~\ref{tab:Sys-BSys-C} shows Sys-B winning on all four metrics under both judges. Win
ratios from per-conversation head-to-head comparisons: Gemini---Sys-B wins 26.5\%, Sys-C
wins 17.0\%, tie 56.5\% (Sys-B wins $1.56\times$ more often); for the Claude judge, Sys-B wins 34.0\%,
Sys-C wins 22.0\%, tie 44.0\% (Sys-B wins $1.55\times$). Differences are $\approx$2.4
standard errors from zero ($p<0.02$); Cohen's $d \approx 0.15$--0.21. 

\begin{table}[t]
\centering
\small
\setlength{\tabcolsep}{4pt}
\begin{tabular}{lcc}
\toprule
\textbf{Metric} & \textbf{$\Delta$ Gemini} & \textbf{$\Delta$ Claude} \\
\midrule
Mission Success      & $+$0.19 & $+$0.16 \\
SRP Relevance        & $+$0.29 & $+$0.16 \\
CHAT Helpfulness     & $+$0.23 & $+$0.45 \\
Intent Understanding & $+$0.32 & $+$0.25 \\
\midrule
Speed (mean conv.)   & \multicolumn{2}{c}{Sys-C $-$13\%} \\
\bottomrule
\end{tabular}
\caption{Sys-B vs.\ Sys-C (Llama~3.3~70B), 200 matched conversations.
Architecture is identical; only the generative LLM differs. $\Delta$ columns show
Sys-B$-$Sys-C advantage per judge. Sys-C is 13\% faster but consistently lower
quality across all metrics and both judges.}
\label{tab:Sys-BSys-C}
\end{table}

The largest gap is CHAT Helpfulness (Claude: $+$0.45). Sys-C generates generic,
parametric responses (e.g., ``The Pulsar X2V2 is known for its lightweight design
and advanced sensor technology...'') while Sys-B grounds responses in actual listing
data (e.g., ``The 4K version supports a 4000\,Hz polling rate; the Pro typically
uses 1000\,Hz, though some bundles include the 4K dongle as an upgrade''). Claude
penalises generic CHAT heavily; Gemini is more lenient. Sys-C also over-applies
CHAT: $\sim$75\% of turns vs.\ Sys-B's $\sim$25\%, this occasionally steers buyers
away from their stated mission toward a different product.

Sys-C's only advantage is speed. It is 13\% faster per conversation. Across all buckets but one, the speed saving
does not offset the quality loss. 

\paragraph{Architectural vs.\ LLM quality are orthogonal.}
Removing the intent-extraction step (Sys-A$\rightarrow$Sys-B) improved quality by
reducing noise. Swapping the backbone LLM (Sys-B$\rightarrow$Sys-C) reduced quality by
reducing generative and reasoning capability. Gains on the two separate dimensions are orthogonal and do not substitute for the other.

\subsection{Judge Philosophy: Gemini vs.\ Claude}
\label{sec:judges}

Both judges receive identical evaluation prompts. Table~\ref{tab:Sys-Bfull} shows a
consistent, large gap in absolute scores across all metrics. The disagreement is
sharpest on CHAT Helpfulness ($+$0.97): Gemini rewards informative, well-structured
explanations; Claude requires those explanations to lead to concrete buyer
actions, such as cart additions, confirmed purchases, or explicit satisfaction signals.
Table~\ref{tab:telldist} shows the full score distribution on this metric: Gemini
awards 5/5 to 74\% of conversations; Claude to only 6\%, a 68-percentage-point gap
despite identical prompts. This is not a calibration difference but a fundamental
disagreement about what CHAT responses are \textit{for}.

\begin{table}[t]
\centering
\small
\setlength{\tabcolsep}{5pt}
\begin{tabular}{lrr}
\toprule
\textbf{Score} & \textbf{Gemini} & \textbf{Claude} \\
\midrule
1 & 3.2\% &  1.3\% \\
2 & 5.0\% & 15.2\% \\
3 & 7.2\% & 35.8\% \\
4 & 10.4\% & 41.9\% \\
5 & \textbf{74.2\%} & \textbf{5.9\%} \\
\bottomrule
\end{tabular}
\caption{CHAT Helpfulness score distributions (600-conversation set). Gemini awards
5/5 to 74\% of conversations; Claude to only 6\%---a 68-percentage-point gap despite
identical prompts and rubrics.}
\label{tab:telldist}
\end{table}

Agreement analysis on the conversation union set confirms the divergence is
structural. Exact agreement on CHAT Helpfulness is only 13\%, versus 57--60\% on
Mission Success and Intent Understanding. The two judges are effectively applying
different criteria to the same dimension. Across all metrics, 29.8\% of
conversations show a $\geq$2-point difference between judges (maximum observed:
3~points on a 5-point scale). When the judges disagree directionally, the asymmetry
is striking: on Mission Success, Gemini scores higher in 90 out of 91 disagreements;
on SRP Relevance, in 20 out of 21. The divergence is not noise around a shared
baseline, but rather a systematic directional bias.

This divergence has measurable consequences. Of 2{,}011 conversations, 838 (41.7\%)
score 5/5 on all metrics under Gemini; only 96 (4.8\%) achieve this under both
judges simultaneously, a 36.9-percentage-point gap. Reasoning style is also
consistent with the philosophical split: Gemini produces concise assessments
($\sim$512 characters) describing process per turn; Claude produces detailed
critiques ($\sim$1{,}087 characters, $2.1\times$ longer) scrutinizing whether the
buyer ultimately accomplished their goal. Because these differences emerge from the
same prompt, they reflect deep architectural or training characteristics of the models underlying the LLM judges.

A particularly consequential finding emerges from the conversation-paired
validation set (Sys-A vs.\ Sys-B, evaluated by both judges): the systematic
Gemini--Claude gap ($\approx$0.5~points) is 50$\times$ larger than the Sys-A vs.\ Sys-B
architecture difference as measured by Claude, and 5$\times$ larger as measured by
Gemini. The choice of evaluator has more impact on reported scores than the
architectural choice being evaluated. This makes judge selection a first-class
methodological decision: all architectural comparisons must use the same judge, and
conclusions drawn under one judge may not transfer to the other.

The methodological implication here is that choosing an LLM judge is equivalent to
choosing a quality criterion. If the target production metric is session depth,
return rate, or interaction satisfaction, Gemini's philosophy is
more aligned; if the target is conversion, cart additions, or task completion,
Claude's philosophy is. Running both judges simultaneously, as we
do here, is not redundancy: the gap between their verdicts directly quantifies the
system's process--outcome trade-off in a way neither judge alone can reveal.
We leave for future work running a human preference study to determine which philosophy better reflects real
user outcomes.

\section{Ethical Considerations}
\label{sec:ethics}

Simulation-based evaluation can surface fairness concerns that production testing cannot easily expose. The failure taxonomy in \S\ref{sec:fulleval} reveals that impatient, strict buyers experience 7.5$\times$ higher failure rates than the overall average: a disparity that would be difficult to detect without controlled simulation across diverse buyer personas. Identifying such gaps offline enables architects to address equity issues before they affect real customers. Our approach to offline simulation enables rapid, low-cost architectural iteration across the full buyer typology before any production commitment.

However, simulation is a laboratory, not a substitute for monitoring real user outcomes after deployment. LLM judges embed specific assumptions about quality that may not reflect all buyer needs, as shown in \S\ref{sec:judges}, Gemini rewards process correctness while Claude demands concrete outcomes, and architectural decisions optimized against one judge may systematically disadvantage buyer types valued by the other. Real users  will also exhibit behavior that synthetic personas cannot fully anticipate. Connecting simulation results to production AB test outcomes is an important direction for future validation.



\section{Discussion}
\label{sec:discussion}

\paragraph{Simulation fidelity.}
To validate the two-agent design, we ran 64 no-responder conversations where
a single LLM generates both sides of the dialogue with access to the same search API. The contrast is stark: single-LLM queries average 24~words vs.\ 8~words in
two-agent mode; action commands (\texttt{click on X}, \texttt{put X in cart})
appear in $<$5\% vs.\ 45\% of turns; natural self-corrections (\textit{``oops ok''})
appear consistently in two-agent conversations and are absent in single-LLM
generation. The single-LLM system generates coherent conversations, but does not optimize for the casual, action-focused patterns of
real shoppers. Hence the necessity of agent independence.
Fine-tuning the buyer LLM on real user interaction data, following the behavioral fidelity approach of \citet{wang2025opera} and \citet{lu2025can}, is a natural direction for future work.

\paragraph{Architectural lessons.}
Two clear lessons emerge from the four experiments. First, intermediate processing
steps should be validated empirically: Sys-A's intent-extraction hypothesis was
theoretically well-motivated but empirically wrong.
Second, architecture and LLM quality are independent dimensions: simplifying
architecture (Sys-A$\rightarrow$Sys-B) improved quality by removing noise; downgrading
the backbone LLM (Sys-B$\rightarrow$Sys-C) reduced quality by reducing generative and
reasoning capability.



\section{Conclusion}
\label{sec:conclusion}

We present a modular two-agent simulation framework that enables controlled,
rapid evaluation of conversational shopping assistant architectures using real
search API integration. Four experiments produced concrete findings: simpler
rolling-window memory can outperform intent extraction in both quality and speed;
full-scale evaluation lays bare concentrated failure modes invisible in small-scale
testing; systematic failure analysis enables rapid targeted iteration; and LLM quality contributes independently of
architecture. An important finding is that Gemini and Claude hold fundamentally
different evaluation philosophies from an identical prompt. This highlights that LLM
judge selection is itself a consequential architectural decision in any evaluation
pipeline. The configurable buyer agent and modular responder are built for continued iteration: as shopping assistant systems mature and buyer behavior evolves, both the scenarios tested and the architectures evaluated can be updated independently. Any e-commerce responder that returns SRP and CHAT response components can be evaluated against the buyer agent without modification to either component, making the framework applicable beyond the specific architectures evaluated here.


\bibliography{custom}

\end{document}